\documentclass[11pt,a4paper]{article}
\usepackage[hyperref]{naaclhlt2019}
\usepackage{times}
\usepackage{latexsym}
\usepackage{graphicx,subcaption}
\usepackage{url}
\usepackage{booktabs,multicol,multirow}
\usepackage{enumitem}
\usepackage{soul}
\aclfinalcopy 

\title{Using Similarity Measures to Select Pretraining Data for NER}

\author{\begin{tabular}{cccc}
Xiang Dai$^{1,2}$ & Sarvnaz Karimi$^{1}$ & \textbf{Ben Hachey$^{2,3}$} & \textbf{Cecile Paris$^{1}$}
\end{tabular}\\
\begin{tabular}{cccc}
\multicolumn{4}{c}{$^{1}$CSIRO Data61, Sydney, Australia}\\
\multicolumn{4}{c}{$^{2}$University of Sydney, Sydney, Australia}\\
\multicolumn{4}{c}{$^{3}$Digital Health CRC, Sydney, Australia}\\
\multicolumn{4}{c}{\tt \{dai.dai,sarvnaz.karimi,cecile.paris\}@csiro.au}\\
\multicolumn{4}{c}{\tt ben.hachey@gmail.com} \\
\end{tabular}
}

\usepackage{amsfonts}

\date{}

\begin{document}
\maketitle
\begin{abstract}
Word vectors and Language Models (LMs) pretrained on a large amount of unlabelled data can dramatically improve various Natural Language Processing (NLP) tasks. 
However, the measure and impact of similarity between pretraining data and target task data are left to intuition. 
We propose three cost-effective measures to quantify different aspects of similarity between source pretraining and target task data. 
We demonstrate that these measures are good predictors of the usefulness of pretrained models for Named Entity Recognition (NER) over 30 data pairs. 
Results also suggest that pretrained LMs are more effective and more predictable than pretrained word vectors, but pretrained word vectors are better when pretraining data is dissimilar.
\end{abstract}

\section{Introduction\label{sec-intro}}
Modern neural architectures for NLP are highly effective when provided a large amount of labelled training data~\citep{Zhang:Zhao:NIPS:2015,Conneau:Schwenk:EACL:2017,Bowman:Angeli:EMNLP:2015}. 
However, a large labelled data set is not always readily accessible due to the high cost of expertise needed for labelling or even due to legal barriers. 
Researchers working on such tasks usually spend a considerable amount of effort and resources on collecting useful external data {\em sources} and investigating how to transfer knowledge to their {\em target} tasks~\citep{Qi:Collobert:CIKM:2009,Kim:Riloff:AMIA:2017}. 
Recent transfer learning techniques make the most of limited labelled data by incorporating word vectors or LMs pretrained on a large amount of unlabelled data. 
This produces dramatic improvements over a range of NLP tasks where appropriate unlabelled data is available~\citep{Peters:Ammar:ACL:2017,Peters:Neumann:NAACL:2018,Akbik:Blythe:COLING:2018,Devlin:Chang:NAACL:2019}.

However, there is still a lack of systematic study on how to select appropriate data to pretrain word vectors or LMs. 
We observe a range of heuristic strategies in the literature: 
(1) collecting a large amount of generic data, e.g., web crawl~\citep{Pennington:Socher:EMNLP:2014,Mikolov:Grave:LREC:2018}; 
(2) selecting data from a similar \emph{field} (the subject matter of the content being discussed), e.g., biology~\citep{Chiu:Crichton:BioNLP:2016,Karimi:Dai:BioNLP:2017}; and, 
(3) selecting data from a similar \emph{tenor} (the participants in the discourse, their relationships to each other, and their purposes), e.g., Twitter, or online forums~\citep{Li:Shah:arXiv:2017,Chronopoulou:Baziotis:NAACL:2019}. 
In all these settings, the decision is based on heuristics and varies according to the individual's experience. 
We also conducted a pilot study that suggests that the practitioner's intuition is to prioritise field over tenor (see Section~\ref{section:survey}).

Our overarching goal is to develop a cost-effective approach that, given a NER data set, nominates the most suitable source data to pretrain word vectors or LMs from several options. Our approach builds on the hypothesis that the more similar the source data is to the target data, the better the pretrained models are, all other aspects (such as source data size) being equal.
We propose using target vocabulary covered rate and language model perplexity to select pretraining data. 
We also introduce a new measure based on the change from word vectors pretrained on source data to word vectors initialized from source data and then trained on target data.
Experiments leverage 30 data pairs from five source and six target NER data sets, each selected to provide a range of fields (i.e., biology, computer science, medications, local business) and tenors (i.e., encyclopedia articles, journal articles, experimental protocols, online reviews).

Our contributions can be summarized as below:
\begin{itemize}[noitemsep,topsep=2pt]
\item We propose methods to quantitatively measure different aspects of similarity between source and target data sets and find that these measures are predictive of the impact of pretraining data on final accuracy. 
To the best of our knowledge, this is the first systematic study to investigate LMs pretrained on various data sources.\footnote{Our pretrained word vectors and LMs are publicly available: \href{https://bit.ly/2O0mOOG}{https://bit.ly/2O0mOOG}, and code at \href{https://github.com/daixiangau/naacl2019-select-pretraining-data-for-ner}{https://github.com/daixiangau/naacl2019-select-pretraining-data-for-ner}.}
\item We find that it is important to consider tenor as well as field when selecting pretraining data, contrary to human intuitions.
\item We show that models pretrained on a modest amount of similar data outperform pretrained models that take weeks to train over very large generic data.
\end{itemize}

\section{Related Work}
\paragraph{Text Similarity} 
Word similarity following the hypothesis that similar words tend to occur in similar contexts~\citep{Harris:Word:1954} is well studied and forms the foundation of neural word embedding architectures. 
\citet{Hill:Reichart:CL:2015} and \citet{Budanitsky:Hirst:CL:2006} evaluate functional similarity (as in \emph{school} versus \emph{college}) and associative similarity (as in \emph{school} versus \emph{teacher}) captured by semantic models, respectively.
\citet{Pavlick:Rastogi:ACL:2015} study sentence-level similarity, using entailment relation, vector embedding and stylistic variation measures. 
\citet{Kusner:Sun:ICML:2015} propose Word Mover's Distance to measure the similarity between documents and evaluate on document classification tasks. 
We extend the study of similarity to corpus-level, and focus on its implication on unsupervised pretraining.

\paragraph{Pretrained Word Vectors}
The effectiveness of pretrained word vectors mainly depends on three factors: source data, training algorithm, and its hyper-parameters. 
\citet{Turian:Ratinov:ACL:2010} and \citet{Levy:Goldberg:TACL:2015} systematically compare count-based distributional models and distributed neural embedding models. 
They find that both models can improve the performance of downstream tasks.
\citet{Chiu:Crichton:BioNLP:2016} identify the most influential hyper-parameters of neural embedding methods. They also investigate the impact of the source data size and find that larger pretraining data do not necessarily produce better word vectors for biomedical NER. 
Our work regarding pretrained word vectors is conducted using skip-gram model with default hyper-parameter setting~\citep{Mikolov:Chen:arXiv:2013}, and our focus is on the impact of similarity between source data and target task data on the effectiveness of pretrained word vectors for NER tasks. 
Our observations are a useful supplement to the literature as a practitioners' guide.

\paragraph{Pretrained Language Models}
\citet{Dai:Le:NIPS:2015} investigate different methods to transfer knowledge to supervised recurrent neural networks. 
They establish that a pretrained recurrent LM can improve the generalization ability of the supervised models. 
They use unlabelled data from Amazon reviews to pretrain the LM and find that it can improve classification accuracy on the Rotten Tomatoes data set.
\citet{Joshi:Dai:SMM4H:2018} empirically showed that, for their vaccination behaviour detection task on twitter data, LMs pretrained on a small amount of movie reviews outperform the ones pretrained on large size of Wikipedia data.
\citet{Peters:Ammar:ACL:2017} successfully inject the information captured by a bidirectional LM into a sequence tagger, and extend this approach to other NLP tasks~\citep{Peters:Neumann:NAACL:2018}. 
Our work is based on~\citep{Peters:Neumann:NAACL:2018} and investigates the impact of pretraining data on the effectiveness of pretrained LMs for downstream NER tasks.

\begin{figure*}[ht]
    \centering
    \includegraphics[width=0.95\textwidth]{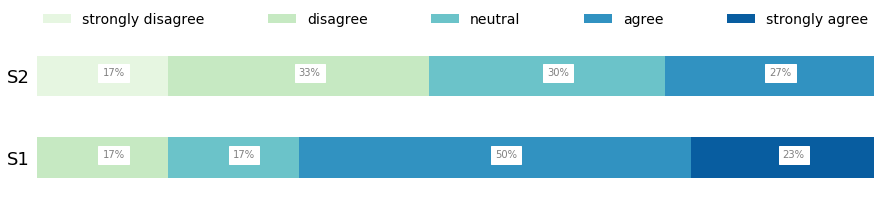}
    \caption{Likert scale ratings from NLP and ML practitioners ($N=30$) for the statement `Unsupervised pretraining on S would be useful for supervised named entity recognition learning on T.' Target data T is described as `Online forum posts about medications,' source data S1 as `Research papers about biology and health,' and source data S2 as `Online reviews about restaurants, hotels, barbers, mechanics, etc.'~\label{fig:human-intuition}}
\end{figure*}

\paragraph{Transfer Learning}
While our study falls into the paradigm of semi-supervised learning, we distinguish ourselves from other studies in transfer learning. 
One sub-area of transfer learning is domain adaptation, which aims to learn transferable representation from a source domain and apply it to a target domain~\citep{Blitzer:McDonald:EMNLP:2006,Yang:Eisenstein:NAACL:2015}. 
The question in domain adaptation is usually framed as `Given a source and a target, how to transfer?'. 
In contrast, the question we address is `Given a specific target, which source to choose from?'. 
The other sub-area of transfer learning is transferring from multiple sources~\citep{Yin:Schutze:CoNLL:2015,Li:Baldwin:NAACL:2018}. 
Our work focuses, instead, on the selection of a single external data source. 
Our work is inspired by the methodology proposed by~\citet{Johnson:Anderson:ACL:2018} where they predict a system's accuracy using larger training data from its performance on much smaller pilot data. 
However, we aim to predict the usefulness of pretrained models for target tasks from the similarity between the source pretraining data and the target task data.

\paragraph{Named Entity Recognition}
Our work builds on the literature on deep neural networks applied to sequence tagging tasks.
Architectures based on different combinations of convolutional and recurrent neural networks have achieved state-of-the-art results on many NER tasks. 
A detailed review and comparison of these methods can be found in~\citep{Yang:Liang:COLING:2018}. 
Our experiments on the usefulness of pretrained word vectors and pretrained LMs for NER tasks are based on one variant proposed by~\citet{Lample:Ballesteros:NAACL:2016}.

\section{What Human Intuition Indicates}
\label{section:survey}
Results of a survey capturing intuition regarding selection of pretraining data across 30 NLP or machine learning practitioners is shown in Figure~\ref{fig:human-intuition}. Participants were provided short descriptions of the target data set T, and two possible source data sets S1 and S2 as
\begin{itemize}
    \item T: Online forum posts about medications;
    \item S1: Research papers about biology and health; 
    \item S2: Online reviews about restaurants, hotels, barbers, mechanics, etc.
\end{itemize}

We constructed each of these descriptions as `$t$ about $f$' where $t$ is intended to describe the tenor and $f$ the field. Each participant rated both sources on a five-point Likert, indicating agreement with the statement ``Unsupervised pretraining on S would be useful for supervised named entity recognition learning on T''.  73\% of participants agreed or strongly agreed that S1 would be useful, while only 27\% agreed that S2 would be useful. A Wilcoxon signed-rank test indicates that scores are significantly higher for S1 than for S2 ($Z = 43.0, p < 0.001$). 

Although small in scale, these results show that intuition varies across practitioners, motivating our work on identifying quantitative measures that are predictive of performance. These results also suggest that practitioners favour field over tenor when selecting pretraining data, which would be detrimental to accuracy of the target NER tasks in later experiments (Section~\ref{sec-impactpretrainingdata}).

\section{Similarity Measures}
To measure the similarity between source and target data, we start from identifying linguistic concepts behind these human intuitions. 
Then, we propose several measures to quantify these attributes which lead to the perception that two data sets are similar. 

Researchers who select pretraining data from a similar field believe that, if the source data has a similar field to the target data, they tend to share similar vocabulary. 
Conversely, vocabularies are different from each other if source and target are from different fields. 
Imagine data sets about medications and restaurants. 
Those who select pretraining data from a similar tenor believe that tenor may impact the writing style of text. 
Imagine the participants in online reviews and scientific papers, their relationships to each other, their purposes and how these affect text style, including punctuation, lexical normalization, politeness, emotiveness and so on~\citep{Lee:LLT:2001,Solano:TCR:2006,Pavlick:Tetreault:TACL:2016}.

Below, we detail different measures based on these intuitions to quantify different aspects of similarity between two data sets.

\subsection{Target Vocabulary Covered}
The first measure is simply the percentage of the target vocabulary that is also present in the source data.
An extremely dissimilar example is that of different languages. 
They have a totally different vocabulary and are considered dissimilar, even if they are written in a similar style and talking about the same subject~\footnote{Our focus is on transferring through pretrained models using one single source and we do not consider multilingual similarity.}.
We propose \emph{Target Vocabulary Covered (TVC)} as a measure of field, calculated as
\[
TVC (D_S, D_T) = \frac{|V_{D_S} \cap V_{D_T}|}{|V_{D_T}|},
\]
where $V_{D_S}$ and $V_{D_T}$ are sets of unique tokens in source and target data sets respectively. 
We also investigate a variant where only content words (nouns, verbs, adjectives) are used to calculate $V_{D_S}$ and $V_{D_T}$. 
We denote this variant as \emph{VCcR}.

\subsection{Language Model Perplexity}
A language model can assign a probability to any sequence of words $<w_1,\cdots,w_N>$ using chain rule of probability:
\[
p(w_1, w_2, \cdots, w_N) = \prod_{i=1}^N p(w_i|w_1^{w-1}),
\]
where $N$ is the length of the sequence and $w_1^{i-1}$ are all words before word $w_i$.
In practice, this equation can be simplified by n-gram models based on Markov Assumption:
\[
p(w_1, w_2, \cdots, w_N) = \prod_{i=1}^N p(w_i|w_{i-n+1}^{i-1}),
\]
where $w_{i-n+1}^{i-1}$ represents only $n$ preceding words of $w_i$.
To make the model generalize better, smoothing techniques can be used to assign non-zero probabilities to unseen events. 
In this study, we use Kneser-Ney smoothed 5-gram models~\citep{Heafield:SMT:2011}. 
To measure the similarity between two data sets using language modeling, we first train the language model on the source data, then evaluate it on the target data using perplexity to represent the degree of similarity. 
The intuition is that, if the model finds a sentence very unlikely (dissimilar from the data where this language model is trained on), it will assign a low probability and therefore high perplexity. 
The summed up \emph{perplexity (PPL)} is then:
\[
PPL(D_S, D_T) = \sum_{i=1}^{m} P(D_T^i)^{-\frac{1}{N_i}},
\]
where $m$ is the number of sentences in the target data set, and $P(D_T^i)$ is the probability assigned by the language model trained on the source data to the $i$-th sentence from the target data set, whose sentence length is $N_i$.

PPL is token-based, similar to TVC, but also captures surface structure. 
We therefore propose PPL as a proxy to measure tenor as well as field.

\subsection{Word Vector Variance}
Pretrained word vectors capture semantic and syntactic regularities of words~\citep{Artetxe:Labaka:CONLL:2018}. 
The variance of a word vector that is first trained on the source data and then on the target data can reflect the difference of linguistic regularities between the two data sets. 

Intuitively, if the context words around a given word are very different in the source and target data, then the word vector of this word learned from the source will be updated more than those words whose context words are similar between source and target. 
Therefore, we use \emph{Word Vector Variance (WVV)} as another combined measure of tenor and field.
 
To calculate word vector variance, we first train word vectors on the source data set using skip-gram model~\citep{Mikolov:Chen:arXiv:2013}. 
The trained word vectors are denoted as $WS \in \mathbb{R}^{|V_S| \times d}$, where $|V_S|$ is the vocabulary size of the source data set and $d$ is the vector dimension. 
Then, we use $WS$ as initial weights of a new skip-gram model, and train this new model on the target data. 
We denote the final word vectors as $WT$. 
The WVV can be calculated as:
\[
WVV(D_S, D_T) = \frac{1}{|V_S|} \frac{1}{d} \sum_i^{|V_S|} \sum_j^d (WS^j_i - WT^j_i)^2.
\]
The smaller the word vector variance, the more similar context surrounds the same words from the two data sets, and therefore the more similar the two data sets are.

\section{Data Sets}
\paragraph{Source data sets}
We use five data sets as source data, covering a range of fields (i.e., clinical, biomedical, local business and Wiki with diverse fields) and tenors (i.e., popular reporting, notes, scholarly publications, online reviews and encyclopedia). 
To isolate the impact of source size, we sample all source data to approximately 100 million tokens. 
We also analyze the impact of source data size separately in Section~\ref{section:size}. 
The specifications of these source data sets are given in Table~\ref{tbl-sourcedata}.

\begin{table}[tb]
\begin{footnotesize}
\begin{center}
\begin{tabular}{p{0.15\linewidth}p{0.7\linewidth}}
\toprule
\bf Data set & \bf Description \\ \midrule
1BWB & The original one billion word language model benchmark data~\citep{Chelba:Mikolov:arXiv:2013}, produced from News Crawl data. It has been randomly shuffled and we use the last 13 out of 100 files. \\ \midrule
MIMIC & A clinical database comprising over 58,000 hospital admissions for intensive care unit (ICU) patients~\citep{Johnson:Pollard:SD:2016}. We use the first 50,000 notes associated with hospital stays. \\ \midrule 
PubMed & Around 30 million citations for biomedical literature covering the fields of biomedical and health. We use articles published after October 2017 and utilize their titles and abstracts only. \\ \midrule 
Wiki & WikiText-103, released by~\citet{Merity:Xiong:arXiv:2016} and consisting of around 28K \emph{Good} or \emph{Featured} articles from Wikipedia. These articles are reviewed by human editors, and they are selected based on the writing quality. We refer to this data set as Wiki. \\ \midrule %
Yelp & An online forum where customers can write reviews about local businesses. We use data released in round 12 of the Yelp Data set Challenge and select the first 2 out of 6 million reviews. \\ \bottomrule
\end{tabular}
\caption{\label{tbl-sourcedata}List of the source data sets.}
\end{center}
\end{footnotesize}
\end{table}

\paragraph{Target data sets}
Six NER data sets are used as target data: CADEC~\cite{Karimi:Metke:JBI:2015}, CoNLL2003~\cite{Sang:Meulder:CONLL:2003}, CRAFT~\cite{Bada:Eckert:BMC:2012}, JNLPBA~\cite{Collier:Kim:BioNLP:2004}, ScienceIE~\cite{Augenstein:Das:SemEval:2017} and WetLab~\cite{Kulkarni:Xu:NAACL:2018}. 
Details of these target data are listed in Table~\ref{tbl-targetdata}. 
We choose these data sets based on two considerations: 

\begin{table*}[tb]
\begin{small}
\begin{center}
\begin{tabular}{p{0.1\linewidth}p{0.09\linewidth}p{0.43\linewidth}p{0.25\linewidth}}
\toprule
\bf Data set & \bf Size & \bf Entity Types & \bf Description \\ \midrule
CADEC & 120,341 & Adverse Drug Event, Disease, Drug, Finding, Symptom & Posts taken from AskaPatient, which is a forum where consumers can discuss their experiences with medications. \\ \midrule
CoNLL2003 & 301,418 & Person, Organization, Location, Miscellany & Newswire from the Reuters RCV1 corpus. \\ \midrule
CRAFT & 561,015 & Cell, Chemical entity, Biological taxa, Protein, Biomacromolecular sequence, Entrez gene, Biological process and molecular function, Cellular component & Full-length, open-access journal articles about biology. \\ \midrule
JNLPBA & 593,590 & Protein, DNA, RNA, Cell line and Cell
type & Abstract of journal articles about biology. \\ \midrule
ScienceIE & 99,555 & Process (including methods, equipment), Task and Material (including corpora, physical materials) & Journal articles about Computer Science, Material Sciences and Physics. \\ \midrule
Wetlab & 220,618 & Action, 9 object-based (Amount, Concentration, Device, Location, Method, Reagent, Speed, Temperature, Time) entity types, 5 measure-based (Numerical, Generic-Measure, Size, pH, Measure-Type) and 3 other (Mention, Modifier, Seal) types & Protocols written by researchers about conducting biology and chemistry experiments. \\ 
\bottomrule
\end{tabular}
\caption{\label{tbl-targetdata}List of the target NER data sets and their specifications. Size is shown in number of tokens.}
\end{center}
\end{small}
\end{table*}

\begin{enumerate}
\item NER is a popular structured NLP task. 
Using NER, we want to observe how the similarity between source and target data may affect the effectiveness of different pretrained word vectors and LMs on downstream tasks.
\item NER is highly sensitive to word representations, because the model needs to make token level decisions. 
That is, each token needs to be assigned a proper label. 
Past studies have shown that removing pretrained word vectors from a tagging system results in a large drop in performance~\cite{Huang:Xu:arXiv:2015,Lample:Ballesteros:NAACL:2016}.
\end{enumerate}

\section{Experimental Setup}
\label{section:experimental-setup}
To investigate the impact of source data on pretrained word vectors and LMs, we pretrain word vectors and LMs on different sources separately, then observe how the effectiveness of these pretrained models varies in different NER data sets.

We use the BiLSTM-CRF model, a state-of-the-art model for sequence tagging tasks, as a supervised model for the target NER task. 
We follow the architecture proposed in~\citep{Lample:Ballesteros:NAACL:2016}, except that we use two BiLSTM-layers and employ a CNN network to learn character-level representations~\citep{Ma:Hovy:ACL:2016}. 
Micro average $F_1$ score is used to evaluate the performance of the tagger~\citep{Sang:Meulder:CONLL:2003}.

Word vectors are pretrained using word2vec with its default hyper-parameter setting~\citep{Mikolov:Chen:arXiv:2013}. 
In different experiments, we only replace the word embedding weights initialized by word vectors pretrained on different source data, then make these weights trained jointly with other model parameters. 
The baseline is denoted as {\em None} in Table~\ref{tab:similarity_values}, where word embedding weights are randomly initialized. 

LMs are pretrained using the architecture proposed by~\citet{Jozefowicz:Vinyals:arXiv:2016} with hyper-parameters in~\citep{Peters:Neumann:NAACL:2018}. 
The supervised model used for NER is the same BiLSTM-CRF model mentioned above, and we follow the approach proposed by~\citet{Peters:Neumann:NAACL:2018} to incorporate the pretrained LMs.
Note that these pretrained LMs are character-based. 
Therefore, words in the target data set are first converted into a sequence of characters, and then fed into the LMs. 
The contextualized representation of each word is generated using the outputs of all layers of the pretrained LMs, then injected to the input of the second BiLSTM layer of the supervised model.

\section{Experimental Results}
Using our proposed similarity measures, we first quantify the similarity between all source-target pairs (Section~\ref{section:similarity_results}), then investigate how these measures can be used to predict the usefulness of pretraining data (Section~\ref{sec-impactpretrainingdata}). Finally, we take the source data size into consideration, and observe its impact on the effectiveness of pretrained model on both similar and dissimilar source-target settings (Section~\ref{section:size}).

\subsection{Similarity Between Source and Target Data Sets}
\label{section:similarity_results}
\begin{table*}[ht!]
\begin{small}
\begin{center}
\setlength{\tabcolsep}{4pt} 
\begin{tabular}{r l | c c c c | c c | c c }
\toprule
& & \multicolumn{4}{c}{\bf Similarity} & \multicolumn{4}{|c}{\bf NER $F_1$ Score} \\ 
\cline{7-10}
& & \multicolumn{4}{c}{} & \multicolumn{2}{|c}{\bf Pretrained word vectors} & \multicolumn{2}{|c}{\bf Pretrained LMs} \\ 
\cline{3-6}\cline{7-10} \noalign{\vskip\arrayrulewidth} 
\bf Target & \bf Source & \bf PPL & \bf WVV & \bf TVC (\%) & \bf TVcC (\%) & \bf $F_1$ score & \bf $\Delta$ & \bf $F_1$ score & \bf $\Delta$ \\ 
\midrule
\multirow{6}{*}{CADEC} & None & -- & -- & -- & -- & 66.14 ($\pm$ 0.53) & -- & 66.14 ($\pm$ 0.53) & -- \\ 
 & 1BWB & \phantom{0}307.4 & 1.137 & \bf 81.73 & \bf 82.94 & 69.44 ($\pm$ 0.52) & 3.30 & 70.08 ($\pm$ 0.43) & 3.94 \\
 & MIMIC & 1007.0 & 1.134 & 78.19 & 81.69 & 69.65 ($\pm$ 0.43) & 3.51 & 70.11 ($\pm$ 0.48) & 3.97\\
 & PubMed & \phantom{0}927.4 & 1.195 & 78.81 & 79.79 & 69.84 ($\pm$ 0.55) & 3.70 & 70.15 ($\pm$ 0.50) & 4.01 \\
 & Wiki & \phantom{0}519.8 & 1.196 & 79.74 & 76.71 & 69.62 ($\pm$ 0.15) & 3.48 & 69.32 ($\pm$ 0.65) & 3.18 \\
 & Yelp & \bf \phantom{0}291.1 & \bf 1.104 & 80.76 & 82.28 & \bf 70.27 ($\pm$ 0.34) & \bf 4.13 & \bf 70.46 ($\pm$ 0.52) & \bf 4.32 \\ 
\midrule
\multirow{6}{*}{CoNLL2003} & None & -- & -- & -- & -- & 82.08 ($\pm$ 0.38) & -- & 82.08 ($\pm$ 0.38) & -- \\ 
 & 1BWB & \bf \phantom{0}480.6 & \bf 1.020 & \bf 75.64 & \bf 87.35 & \bf 86.36 ($\pm$ 0.29) & \bf 4.28 & \bf 89.78 ($\pm$ 0.12) & \bf 7.70 \\
 & MIMIC & 2945.0 & 1.542 & 34.47 & 39.55 & 84.94 ($\pm$ 0.35) & 2.86 & 83.68 ($\pm$ 0.30) & 1.60 \\
 & PubMed & 3143.1 & 1.356 & 53.29 & 68.41 & 85.56 ($\pm$ 0.46) & 3.48 & 84.15 ($\pm$ 0.22) & 2.07 \\
 & Wiki & \phantom{0}650.4 & 1.159 & 66.21 & 80.87 & 86.32 ($\pm$ 0.28) & 4.24 & 89.11 ($\pm$ 0.23) & 7.03 \\
 & Yelp & 2025.5 & 1.399 & 53.92 & 68.95 & 85.58 ($\pm$ 0.26) & 3.50 & 85.19 ($\pm$ 0.38) & 3.11 \\
\midrule
\multirow{6}{*}{CRAFT} & None & -- & -- & -- & -- & 69.17 ($\pm$ 0.64) & -- & 69.17 ($\pm$ 0.64) & -- \\ 
 & 1BWB & 1328.1 & 2.073 & 59.07 & 62.98 & 73.97 ($\pm$ 0.06) & 4.80 & 71.23 ($\pm$ 0.81) & 2.06 \\
 & MIMIC & 2427.5 & 2.390 & 48.73 & 50.03 & 73.01 ($\pm$ 0.22) & 3.84 & 71.90 ($\pm$ 0.26) & 2.73 \\
 & PubMed & \bf \phantom{0}360.3 & \bf 1.838 & \bf 76.29 & \bf 80.69 & \bf 75.45 ($\pm$ 0.28) & \bf 6.28 & \bf 75.45 ($\pm$ 0.09) & \bf 6.28 \\
 & Wiki & \phantom{0}974.7 & 2.075 & 63.66 & 63.12 & 74.07 ($\pm$ 0.40) & 4.90 & 69.75 ($\pm$ 0.09) & 0.58 \\
 & Yelp & 2085.7 & 2.187 & 48.01 & 50.85 & 72.48 ($\pm$ 0.13) & 3.31 & 72.75 ($\pm$ 0.26) & 3.58 \\
\midrule
\multirow{6}{*}{JNLPBA} & None & -- & -- & -- & -- & 70.45 ($\pm$ 0.21) & -- & 70.45 ($\pm$ 0.21) & -- \\ 
 & 1BWB & 1190.8 & 2.000 & 39.90 & 53.54 & 72.39 ($\pm$ 0.23) & 1.94 & 72.54 ($\pm$ 0.34) & 2.09 \\
 & MIMIC & 2533.4 & 2.172 & 36.95 & 50.04 & \bf 73.24 ($\pm$ 0.29) & \bf 2.79 & 71.76 ($\pm$ 0.13) & 1.31 \\
 & PubMed & \bf \phantom{0}205.9 & \bf 1.597 & \bf 58.87 & \bf 80.17 & 72.77 ($\pm$ 0.65) & 2.32 & \bf 74.29 ($\pm$ 0.40) & \bf 3.84 \\
 & Wiki & \phantom{0}717.9 & 2.036 & 42.34 & 53.05 & 72.77 ($\pm$ 0.27) & 2.32 & 72.42 ($\pm$ 0.23) & 1.97 \\
 & Yelp & 2134.4 & 2.155 & 30.78 & 41.41 & 72.53 ($\pm$ 0.18) & 2.08 & 72.51 ($\pm$ 0.21) & 2.06 \\
\midrule
\multirow{6}{*}{ScienceIE} & None & -- & -- & -- & -- & 26.85 ($\pm$ 0.17) & -- & 26.85 ($\pm$ 0.17) & -- \\ 
 & 1BWB & \phantom{0}884.6 & 1.197 & 71.50 & 76.78 & 34.40 ($\pm$ 0.50) & 7.55 & 38.10 ($\pm$ 0.31) & 11.25 \\
 & MIMIC & 2706.7 & 1.461 & 54.29 & 59.34 & 31.23 ($\pm$ 0.15) & 4.38 & 35.27 ($\pm$ 0.43) & 8.42 \\
 & PubMed & \bf \phantom{0}345.6 & \bf 1.037 & \bf 83.25 & \bf 87.01 & \bf 37.91 ($\pm$ 0.12) & \bf 11.06 & \bf 42.07 ($\pm$ 0.03) & \bf 15.22 \\
 & Wiki & \phantom{0}684.2 & 1.127 & 76.99 & 78.01 & 36.15 ($\pm$ 0.11) & 9.30 & 40.39 ($\pm$ 0.05) & 13.54 \\
 & Yelp & 1562.2 & 1.347 & 62.32 & 66.42 & 33.92 ($\pm$ 0.14) & 7.07 & 36.05 ($\pm$ 0.02) & 9.20 \\
\midrule
\multirow{6}{*}{WetLab} & None & -- & -- & -- & -- & 76.91 ($\pm$ 0.10) & -- & 76.91 ($\pm$ 0.10) & -- \\ 
 & 1BWB & 1526.0 & 2.167 & 59.67 & 61.47 & 78.66 ($\pm$ 0.35) & 1.75 & 78.94 ($\pm$ 0.05) & 2.26 \\
 & MIMIC & 3046.1 & 2.393 & 53.83 & 55.31 & 78.68 ($\pm$ 0.14) & 1.13 & 78.65 ($\pm$ 0.13) & 1.74 \\
 & PubMed & \bf 1104.7 & \bf 2.078 & \bf 71.39 & \bf 74.46 & \bf 78.93 ($\pm$ 0.28) & \bf 2.02 & \bf 79.62 ($\pm$ 0.07) & \bf 2.71 \\
 & Wiki & 1617.8 & 2.158 & 61.02 & 60.31 & 78.45 ($\pm$ 0.20) & 1.54 & 79.05 ($\pm$ 0.21) & 2.14 \\
 & Yelp & 1784.5 & 2.240 & 54.16 & 54.96 & 78.48 ($\pm$ 0.15) & 1.57 & 79.04 ($\pm$ 0.19) & 2.13 \\
\bottomrule
\end{tabular}
\end{center}
\end{small}
\caption{\label{tab:similarity_values} Similarity between source and target data sets (left), and the effectiveness of word vectors and LMs pretrained using different sources for NER (right). Lower PPL or WVV values indicate higher similarity between source and target, while higher TVC and TVcC values indicate higher similarity. \emph{None} rows refer to the models that word embedding weights are randomly initialized with no pretrained LMs. \emph{$\Delta$} shows absolute improvement. We repeat every NER experiment 5 times, and report mean and standard deviation of test $F_1$ scores.}
\end{table*}%

Different aspects of similarity measured between five source and six target data sets are shown in the left side of Table~\ref{tab:similarity_values}. 
The language model trained on PubMed achieves lower perplexity when evaluated on CRAFT, JNLPBA and ScienceIE compared to other sources. On one hand, it is expected that PubMed is similar to CRAFT and JNLPBA, since they are all sampled journal articles about biology and health, thus being similar in terms of both field and tenor. On the other hand, although ScienceIE does not have the same field as PubMed (computer science, material and physics versus biology and health), they are similar because they share a similar tenor (scholarly publications).

The measures calculated on CADEC also show that tenor is reflected more than field by PPL and WVV. 
Source data set Yelp is more similar to CADEC than PubMed and MIMIC from both PPL and WVV perspectives. 
CADEC is a data set focusing on recognizing drugs, diseases and adverse drug events. 
The field of CADEC is therefore more similar to PubMed which includes journal articles in health discipline and MIMIC which contains clinical notes. However, CADEC is written by patients, and can be considered as `drug reviews'. The tenor is therefore closer to the one in Yelp, where customers use informal language to describe their experiences. 

All sources are measured against WetLab with relatively high PPL and WVV values. 
This reflects the fact that the tenor of WetLab (experimental protocols) is different from the tenor of all sources, although WetLab has a similar field (biology) with PubMed which is therefore more similar than other sources. 
For CoNLL2003, 1BWB which is News Crawl data is the most similar source, while PubMed is the most dissimilar source from PPL perspective, and MIMIC is the most dissimilar one using WVV measure. 

Although WVV does not distinguish between different sources as PPL does, it still reflects the same trend as PPL regarding which source is the most similar to a given target data set.

\paragraph{Can these different similarity measures reach a consensus?}
Similarity results in Table~\ref{tab:similarity_values} indicate that using different measures can lead to almost the same answer regarding which source is the most similar one to a given target. 
To further investigate the level of agreement between different similarity measures, we employ inter-method agreement that we ask a fine-grained question on the results in Table~\ref{tab:similarity_values}: given a target and two sources, do similarity measures make the same conclusion as to which source is more similar? 
Using the five source and six target data sets, we generate a total of 60 binary comparisons.
For example, given WetLab, is 1BWB a more similar source than Wiki? 
PPL shows that 1BWB is more similar, while WVV gives an opposite answer. 
Fleiss's kappa~\cite{Fleiss:1971} (a variant of Cohen's kappa for more than two raters) is a robust metric used to measures inter-rater agreement, since it takes random chance into consideration. 
We use it to measure the inter-method agreement between the 60 binary comparisons inferred using PPL, WVV and TVC. 
Our results achieve a Fleiss's kappa of 0.733, which shows a high agreement between conclusions inferred using different measures.

Overall, we find that these similarity measures can reach high level of consensus. 
To simplify our following discussion, from here on {\em similar} means low PPL (because of its clear distinction between different sources), unless otherwise stated.

\subsection{Impact of Pretraining Data}\label{sec-impactpretrainingdata}
After we quantify the similarity between source and target data sets, the next question is how these similarity measures can be used to predict the effectiveness of pretrained models for NER tasks. 

Results in Table~\ref{tab:similarity_values} show that, although all pretrained word vectors and LMs can improve the performance of the target model, the improvement varies in different target data sets. 
In other words, no single source is suitable for all target NER data sets. 
Word vectors and LMs pretrained on a source similar to the target outperform the ones pretrained on other sources (except pretrained word vectors for JNLPBA data set). 

We also observe that pretrained LMs provide more benefits than pretrained word vectors if source data is similar to the target (see 1BWB-CoNLL2003 and PubMed-JNLPBA data pairs).
However, if the source is dissimilar to the target, pretrained word vectors outperform pretrained LMs (see these pairs: MIMIC-CoNLL2003, PubMed-CoNLL2003, MIMIC-CRAFT).

\paragraph{Predictiveness of similarity measures}
To analyze how proposed similarity measures correlate to the effectiveness of pretrained word vectors and LMs for the target NER tasks, we employ the Pearson correlation analysis to find out the relationships between improvement due to pretrained models and TVC, TVcC, PPL and WVV. 
The results in Table~\ref{tab:predictiveness} show that our proposed similarity measures are predictive of the effectiveness of the pretraining data. 
In terms of pretrained word vectors, VCcR is the most informative factor in predicting the effectiveness of pretrained word vectors given a target data set. 
It implies that finding a source data set which has large vocabulary intersection with the target data set is a promising first step to generate effective pretrained word vectors. 
The results regarding the LM performance show that it has a stronger correlation with similarity measures than the one of word vectors, thus more predictable using our proposed measures.

\begin{table}[tb]
    \centering
    \begin{small}
    \begin{tabular}{r|c|c}
    \toprule
     & \bf Word vectors & \bf LMs \\ \midrule
     TVC & \phantom{0}0.454 & \phantom{0}0.666 \\
     TVcC & \phantom{0}0.469 & \phantom{0}0.739 \\
     PPL & -0.398 & -0.618 \\
     WVV & -0.406 & -0.747 \\
     \bottomrule
     \end{tabular}
    \caption{\label{tab:predictiveness}Correlation coefficients between similarity measures and the effectiveness of pretrained models. The coefficients vary between -1 (negative correlation) and 1 (positive correlation). Zero means no correlation.}
    \end{small}
\end{table}%

\paragraph{Comparison to publicly available pretrained models}
Recent literature shows substantial improvements are sometimes possible when pretraining on very large generic corpora. 
Given that pretrained models are freely available, is it even necessary to pretrain on similar data as proposed above? 
We compare to publicly available (1) word vectors trained on 6 billion tokens of encyclopaedia articles and news stories about various fields~\footnote{\href{https://nlp.stanford.edu/projects/glove/}{https://nlp.stanford.edu/projects/glove/}} and (2) LMs trained on 5.5 billion tokens of encyclopaedia articles and news stories about various fields~\footnote{\href{https://allennlp.org/elmo}{https://allennlp.org/elmo}}. We use the same experimental setup described in Section~\ref{section:experimental-setup}, that pretrained word vectors are used to initialize the weights of word embedding layer, whereas outputs of pretrained LMs are used as input features of the supervised model.
We find that word vectors and LMs pretrained on small similar sources can achieve competitive or even better performance than the ones pretrained on larger sources (Table~\ref{tab:compare-public}). 
On JNLPBA, ScienceIE and Wetlab, LMs pretrained on the small similar source perform better, while word vectors pretrained on the small similar source perform better on CRAFT, JNLPBA, and ScienceIE.

\begin{table}[tb]
    \centering
    \begin{small}
    \begin{tabular}{r|cc|cc}
    \toprule
     & \multicolumn{2}{c}{\bf Word vectors} & \multicolumn{2}{|c}{\bf LMs} \\
     & GloVe & Ours & ELMo & Ours \\ \midrule
     CADEC & \bf 70.30 & 70.27 & \bf 71.91 & 70.46 \\
     CoNLL2003 & \bf 90.25 & 86.36 & \bf 91.34 & 89.78 \\
     CRAFT & 74.22 & \bf 75.45 & \bf 75.77 & 75.45 \\
     JNLPBA & 73.19 & \bf 73.24 & 73.65 & \bf 74.29 \\
     ScienceIE & 37.10 & \bf 37.91 & 41.15 & \bf 42.07 \\
     WetLab & \bf 79.15 & 78.93 & 79.57 & \bf 79.62 \\
     \bottomrule
     \end{tabular}
     \caption{\label{tab:compare-public}Comparison between our best performance pretrained models and the publicly available ones, which are pretrained on much larger corpora.}
    \end{small}
\end{table}%

These results indicate that a small similar source reduces the computational cost without sacrificing the performance. This is especially important in practice, because collecting data and pretraining models are expensive. For example, a LM pretrained on 1 billion tokens takes three weeks to train on 32 GPUs~\citep{Jozefowicz:Vinyals:arXiv:2016}. 

\begin{table}[pt!]
    \centering
    \begin{small}
    \begin{tabular}{r|c|c|c|c}
    \toprule
    & \multicolumn{2}{c}{\bf ScienceIE} & \multicolumn{2}{c}{\bf WetLab} \\ \hline
    \bf & \bf Def & \bf Opt & \bf Def & \bf Opt \\ \hline
    1BWB & 34.40 & 34.57 & 78.66 & 79.12 \\
    MIMIC & 31.23 & 34.14 & 78.68 & 78.65 \\
    PubMed & \bf 37.91 & \bf 38.86 & \bf 78.93 & \bf 79.28 \\
    Wiki & 36.15 & 35.63 & 78.45 & 78.99 \\
    Yelp & 33.92 & 34.25 & 78.48 & 78.78 \\
    \bottomrule
    \end{tabular}
    \caption{Impact of hyper-parameter setting on the effectiveness of pretrained word vectors. `Opt' is hyper-parameter setting proposed in~\citep{Chiu:Crichton:BioNLP:2016}, whereas `Def' is the default setting in word2vec.}
    \label{tab:best-hyper-parameter}
    \end{small}
\end{table}

\paragraph{Comparison to other hyper-parameter settings}
\citet{Chiu:Crichton:BioNLP:2016} propose a hyper-parameter combination of skip-gram model that is empirically identified on NER tasks. 
They find that a narrow context window size can boost the performance since it can capture better word function rather than domain similarity. 
We use their proposed hyper-parameter setting to train word vectors on different source data, and evaluate these pretrained word vectors on the ScienceIE and WetLab data sets. 
The reason for hand-picking these two is that benefits of pretrained word vectors on these two sets vary with a large margin. 
Our results suggest that this hyper-parameter setting can overall (except Wiki-ScienceIE and MIMIC-WetLab pairs) produce better performance compare to the default setting (Table~\ref{tab:best-hyper-parameter}). 
Most importantly we observe that our observation that similar sources generate better pretrained models can still holds with these hyper-parameters: PubMed, which is the most similar source to both target data sets, still outperforms other sources.

\subsection{Controlling for source data size\label{section:size}}
\begin{figure}[pt!]
    \centering
    \includegraphics[width=0.48\textwidth]{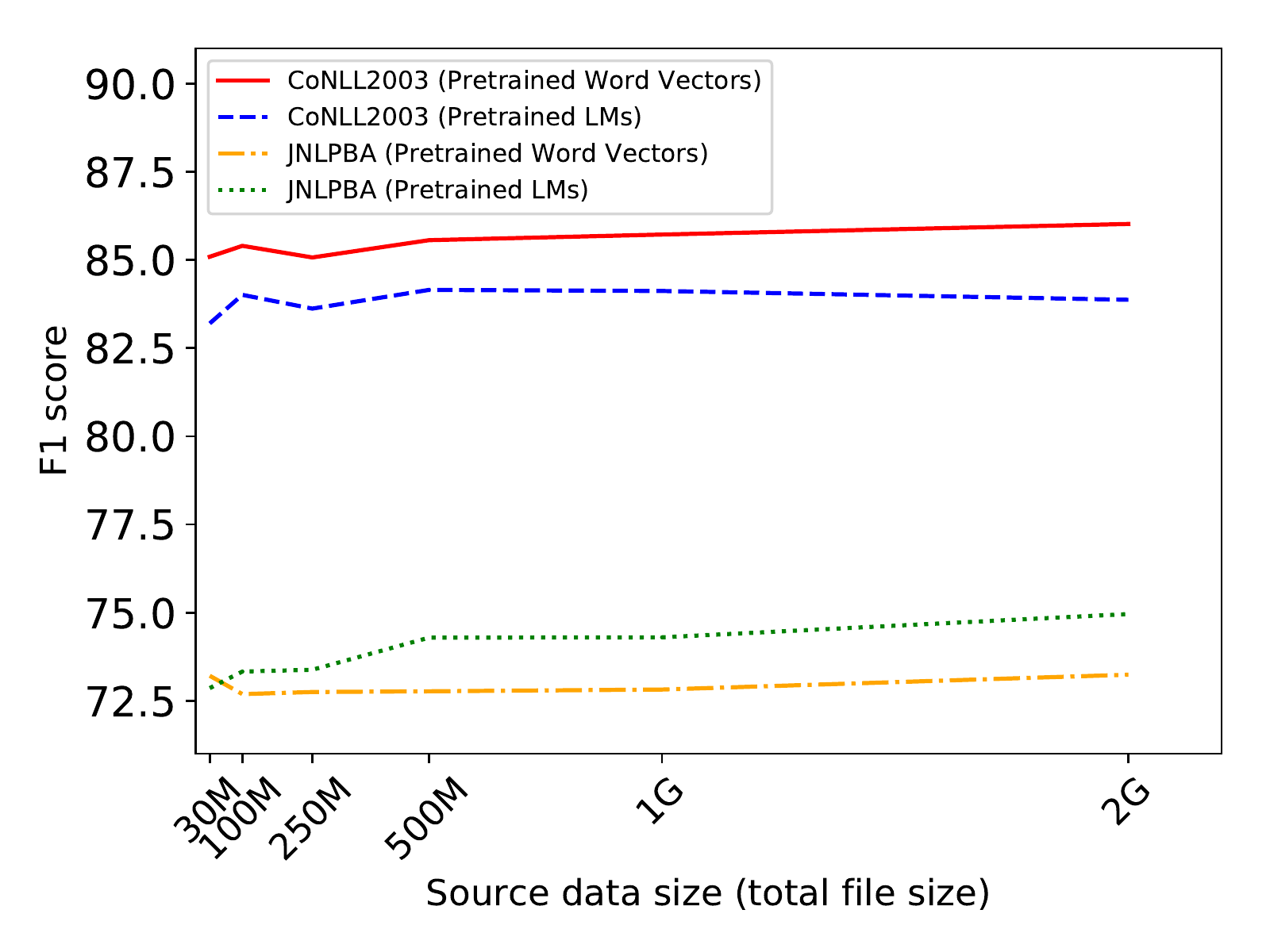}
    \caption{Impact of source data size on the effectiveness of pretrained models for NER.}\label{fig:impact-of-size}
\end{figure}

To further investigate how source data size affects pretrained word vectors and LMs for NER tasks, we sample six PubMed subsets of different size. 
For target data sets, we use CoNLL2003, to which PubMed is the most dissimilar source, and JNLPBA, to which PubMed is the most similar source. 
We observe that 500 MB of pretraining data appears to be sufficient to calculate similarity, and capping factors out the impact of size (Figure~\ref{fig:impact-of-size}). 
As discussed, VCcR is the most influential factor affecting the usefulness of pretrained word vectors for NER task. 
Increasing source data size may provide a larger vocabulary intersection with the target data set, but the resulting absolute $F_1$ score increase is less than 0.5, after the source data has been large enough. 
We also observe that if source and target data are dissimilar (PubMed-CoNLL2003 pair), pretrained word vectors is a better option than pretrained LMs, no matter how large source data is. 
However, pretrained LMs outperform pretrained word vectors, if source is similar to target (PubMed-JNLPBA pair). 

We leave exploration of the combined effect of size and similarity to future work, but believe size should be considered separately, noting that results here suggest that similarity is more important.

\section{Conclusions}
We studied whether there are cost-effective methods to identify data sets to pretrain word vectors and LMs that are building blocks of NER models. 
We proposed using three measures, Target Vocabulary Covered, Language Model Perplexity, and Word Vector Variance, to measure different aspects of similarity between source and target data. 
We investigated how these measures correlate with the effectiveness of pretrained word vectors and LMs for NER tasks. 
We found that the effectiveness of pretrained word vectors strongly depends on whether the source data have a high vocabulary intersection with target data, while pretrained LMs can gain more benefits from a similar source. 
While different NLP tasks may rely on different aspects of language, our study is a step towards systematically guiding researchers on their choice of data for pretraining. 
As a future study, we will explore how these similarity measures predict performance of pretrained models in other NLP tasks.

\section*{Acknowledgments} 
We would like to thank Massimo Piccardi and Mark Dras for their constructive feedback. The authors also thank the members of CSIRO Data61's Language and Social Computing (LASC) team for helpful discussions, as well as anonymous reviewers for their insightful comments.

\bibliography{naaclhlt2019}
\bibliographystyle{acl_natbib}

\end{document}